\def\year{2021}\relax
\documentclass[letterpaper]{article} % DO NOT CHANGE THIS
\usepackage{aaai21} % DO NOT CHANGE THIS
\usepackage{times} % DO NOT CHANGE THIS
\usepackage{helvet} % DO NOT CHANGE THIS
\usepackage{courier} % DO NOT CHANGE THIS
\usepackage[hyphens]{url} % DO NOT CHANGE THIS
\usepackage{graphicx} % DO NOT CHANGE THIS
\usepackage{graphicx}  % DO NOT CHANGE THIS
\usepackage{natbib} % DO NOT CHANGE THIS OR ADD OPTIONS
\usepackage{caption} % DO NOT CHANGE THIS OR ADD OPTIONS
\usepackage{pifont}
\newcommand{\cmark}
{\ding{51}}%
\newcommand{\xmark}{\ding{55}}%
\usepackage{color}
\begin{document}

\pdfinfo{
/Title (Natural Language Inference in Context - Investigating Contextual Reasoning over Long Texts)
/Author (Hanmeng Liu, Leyang Cui, Jian Liu, Yue Zhang)
/TemplateVersion (2021.1)
}

\title{Natural Language Inference in Context\\ --- Investigating Contextual Reasoning over Long Texts}
\author{Hanmeng Liu,\textsuperscript{\rm 1} Leyang Cui,\textsuperscript{\rm, 1} Jian Liu,\textsuperscript{\rm 2} Yue Zhang\textsuperscript{\rm 3}\\
}
\affiliations{\textsuperscript{\rm 1} Zhejiang University, \textsuperscript{\rm 2} Fudan University, \textsuperscript{\rm 3} Westlake University\\
liuhanmeng@zju.edu.cn,
cuileyang@westlake.edu.cn,
jianliu17@fudan.edu.cn,
yue.zhang@wias.org.cn

}
\maketitle
\begin{abstract}
Natural language inference (NLI) is a fundamental NLP task, investigating the entailment relationship between two texts. Popular NLI datasets present the task at sentence-level. While adequate for testing semantic representations, they fall short for testing contextual reasoning over long texts, which is a natural part of the human inference process. We introduce ConTRoL, a new dataset for \textbf{ConT}extual \textbf{R}easoning \textbf{o}ver \textbf{L}ong Texts. Consisting of 8,325 expert-designed ``context-hypothesis" pairs with gold labels, ConTRoL is a passage-level NLI dataset with a focus on complex contextual reasoning types such as logical reasoning. It is derived from competitive selection and recruitment test (verbal reasoning test) for police recruitment, with expert level quality. Compared with previous NLI benchmarks, the materials in ConTRoL are much more challenging, involving a range of reasoning types. Empirical results show that state-of-the-art language models perform by far worse than educated humans. Our dataset can also serve as a testing-set for downstream tasks like Checking Factual Correctness of Summaries. 
\end{abstract}

\section{Introduction}

Natural languages are powerful tools for reasoning. In NLP, natural language inference (NLI) has attracted surging research interests \cite{snli:emnlp2015, williams-etal-2018-broad, ch2019abductive}. The task is to determine whether a hypothesis \textit{h} can reasonably be inferred from a premise \textit{p}. 
Thanks to the generalizability of the NLI framework (i.e., nearly all questions about meaningfulness in language can be reduced to questions of entailment and contradiction in context),  NLI can serve as a proxy to general tasks such as natural language understanding (NLU). As a result, the NLI task is constantly employed as a testing ground for learning sentence representation as well as evaluating language models, with the expectation of benefiting downstream applications. 

Large-scale NLI datasets have been collected via crowdsourcing. Existing benchmarks \cite{snli:emnlp2015, williams-etal-2018-broad, 10.1007/11736790_9,scitail} handle the task at the sentence-level, generating labelled sentence pairs by probing into the essence of lexical and compositional semantics. These benchmarks explore rich features of sentence meaning, testing various aspects of semantic representation. With the advance of contextualized embeddings such as BERT \cite{devlin-etal-2019-bert}, pre-trained language models achieve competitive results. The state-of-the-art models can even reach human-level performance.

\begin{figure}[t]
\centering
\includegraphics[width=0.45\textwidth]{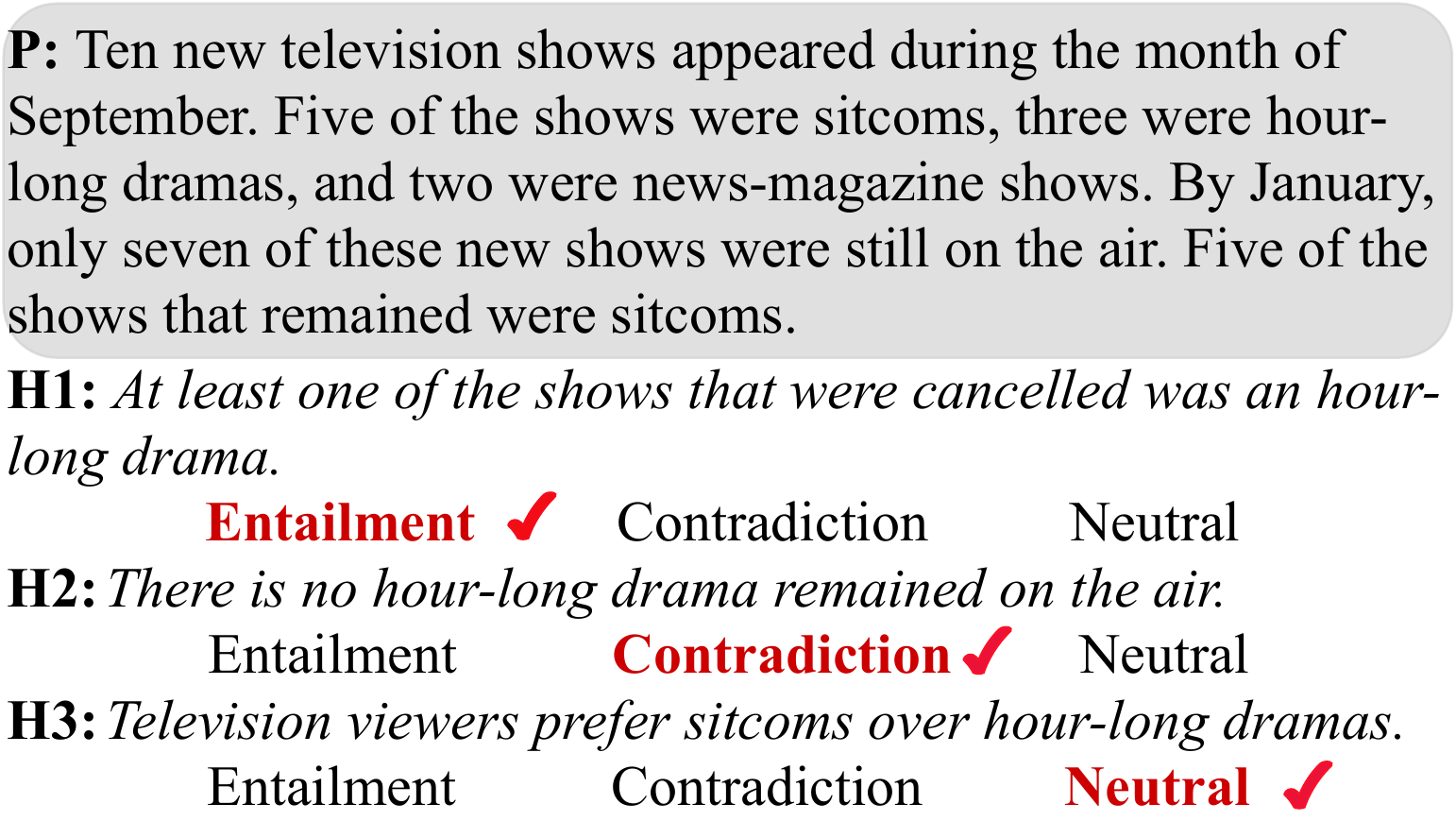} % Reduce the figure size so that it is slightly narrower than the column.
\caption{An example of the ConTRoL dataset. ({\color{red} \cmark} indicates the correct answer.)}
\label{fig:intro}
\end{figure}

%Despite the significant success on the NLI task, language models still struggle to model higher dimensions of meaning beyond sentence-level semantic representations. 
\begin{table*}
\centering
\begin{tabular}{c|cccc}
\hline
\textbf{Dataset} & \textbf{Task} & \textbf{Reasoning} & \textbf{Context} & \textbf{Source} \\
\hline
SQuAD \cite{rajpurkar2016squad} & Reading Comprehension & \cmark & Passage & Wikipedia  \\
\textsc{WIKIHOP} \cite{welbl2017constructing} & Reading Comprehension & \cmark & Document & Wikipedia  \\
\textsc{HotpotQA} \cite{Yang_2018} & Reading Comprehension & \cmark & Document & Wikipedia  \\
%RACE\cite{lai-etal-2017-race} & Reading Comprehension &  \cmark & Passage & Diverse & \cmark\\
% CLOTH\cite{xie2017largescale} & Cloze & \cmark & Passage & Diverse\\
%CoQA \cite{reddy2018coqa} & Conversational QA &  \xmark & Passage & Diverse \\
%SWAG\cite{zellers2018swag} & Plausible Inference &  \cmark & Sentence & Movie & \xmark\\
Cosmos QA \cite{cosmos} & Reading Comprehension &  \cmark & Passage & Webblog \\
Social IQA \cite{social-iqa} & Reading Comprehension & \xmark & Sentence & Social \\
\textsc{WinoGrande} \cite{winogrande} & Coreference Resolution & \xmark & Sentence & Diverse \\
%DREAM \cite{dream} & Reading Comprehension & \xmark & Dialogue & Diverse \\
CommonsenseQA \cite{commonsenseqa} & Reading Comprehension & \xmark & Sentence & Diverse \\
MuTual \cite{mutual} & Next Utterance Prediction & \cmark & Dialogue & Exam \\
ReClor \cite{yu2020reclor} & Reading Comprehension & \cmark & Passage & Exam \\
LogiQA \cite{Liu_2020} & Reading Comprehension & \cmark & Passage & Exam \\
\hline
RTE \cite{inproceedings} & Natural Language Inference &  \xmark & Sentence & Diverse  \\
SNLI \cite{snli:emnlp2015} & Natural Language Inference &  \xmark & Sentence & Captioning  \\
WNLI \cite{DBLP:journals/corr/abs-1804-07461} & Natural Language Inference & \xmark & Sentence & Fiction  \\
QNLI \cite{DBLP:journals/corr/abs-1804-07461} & Natural Language Inference & \xmark & Sentence & Wikipedia  \\
MultiNLI \cite{williams-etal-2018-broad} & Natural Language Inference &  \xmark & Sentence & Diverse \\
Dialogue NLI \cite{welleck2018dialogue} & Natural Language Inference &  \xmark & Sentence & Persona-chat \\
SciTaiL \cite{scitail} & Natural Language Inference &  \xmark & Sentence & Science \\
Adversarial NLI \cite{Nie2019AdversarialNA} & Natural Language Inference & \xmark & Paragraph & Diverse  \\
AlphaNLI \cite{ch2019abductive} & Natural Language Inference &  \xmark & Sentence & Diverse \\
\hline
ConTRoL & Natural Language Inference & \cmark & Passage & Exam \\
\hline
\end{tabular}
\caption{Comparison between our dataset and existing benchmarks. ``Reasoning" refers to contextual reasoning.}
\label{Tab:benchmarks}
\end{table*}

Contextual reasoning is essential to the process of human cognition, where inference is made based on contextual information and a collection of facts \cite{Giunchiglia92contextualreasoning}. Inferring hidden facts from context is an indispensable element of human language understanding. Contextual reasoning is typically preformed on the passage level, where multiple steps may be necessary for inferring facts from given evidences. 
It has been investigated by NLP tasks such as machine reading \cite{lai-etal-2017-race,dream}, retrieval-based dialogue \cite{smn}. However, dominant NLI benchmarks \cite{snli:emnlp2015,williams-etal-2018-broad} investigate the relationship of two sentences, with relatively less attention being payed to the exploration of grounded logical inference \cite{ch2019abductive,clark2020transformers}. 

\begin{figure*}[t]
\centering
\includegraphics[width=\textwidth]{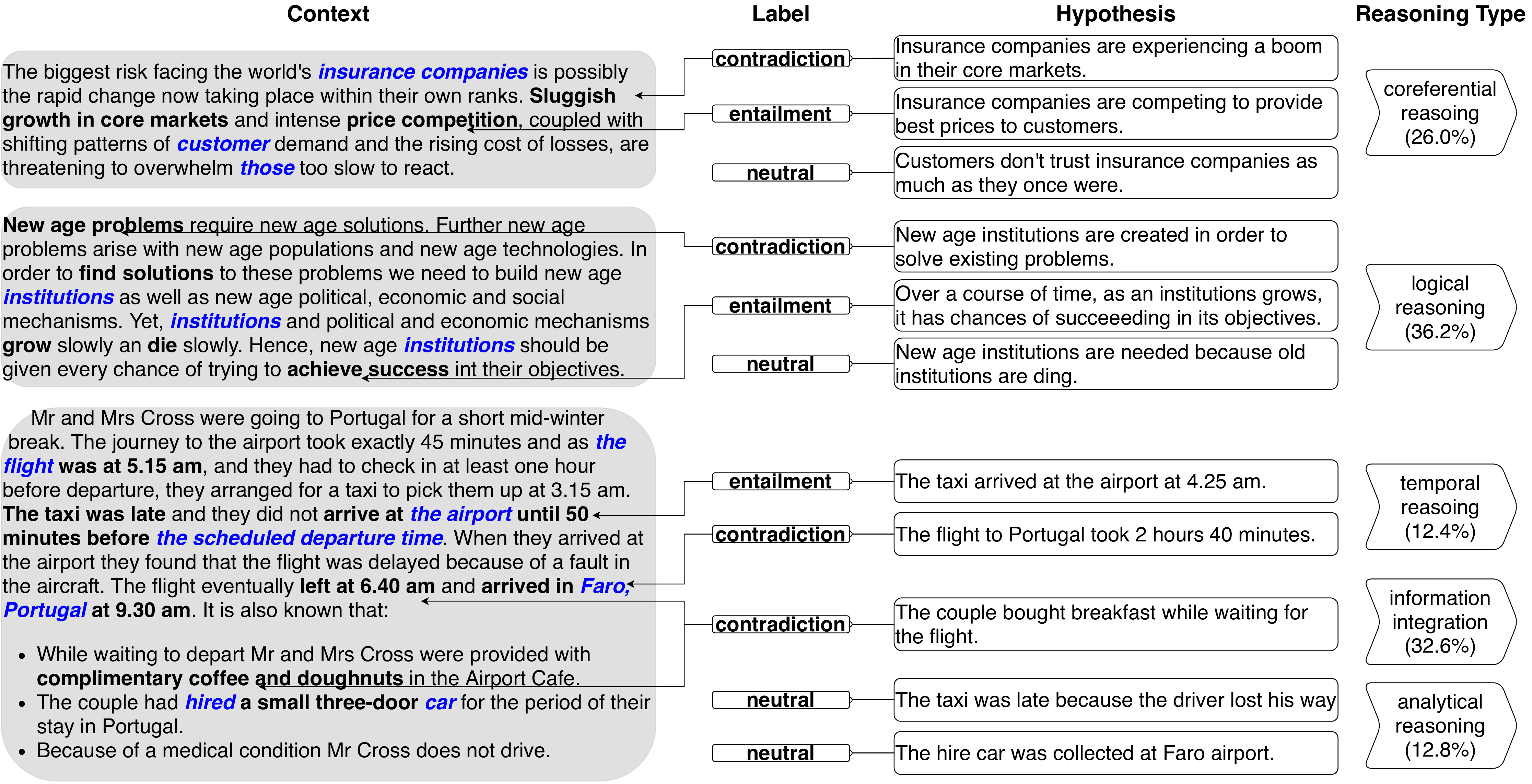} % Reduce the figure size so that it is slightly narrower than the column.
\caption{Reasoning types in ConTRoL (Reasoning clues are highlighted in the context).}
\label{fig:types}
\end{figure*}

We investigate contextual reasoning for NLI by making a dataset that consists of 8,325 instances. One example is shown in Figure \ref{fig:intro}. In this example, the premise consists of several facts concerning a set of shows, which can serve as a context for evidence integration  and reasoning. The truthfulness of the hypotheses are determined by reasoning over multiple sentences. Various types of contextual reasoning are considered in the dataset, with more examples being shown in Figure \ref{fig:types} We name our open-domain dataset \textbf{ConT}extual \textbf{R}easoning \textbf{o}ver \textbf{L}ong Texts (ConTRoL), which is a passage-level natural language inference dataset with gold label data. It differs from the existing NLI datasets in the following three main aspects: (1) the materials are sourced from verbal reasoning exams which are expert-designed rather than crowdsouced; (2) they inspect the abilities of various reasoning types; (3) the contexts are more complex than previous datasets with longer spans.

We evaluate the state-of-the-art NLI models to establish baseline performances for ConTRoL. Experimental results demonstrate a significant gap between machine and human ceiling performance. Detailed analysis is given to shed light on future research. Our dataset and results are released at https://anonymous.

\section{Related Work}

\textbf{Natural Language Inference} The task of text entailment was introduced in the PASCAL Recognizing Textual Entailment (RTE) challenges \cite{10.1007/11736790_9}, which deals with relationship of sentence pairs. On the third RTE challenge \cite{10.5555/1654536.1654538}, a very limited number of longer texts with multiple sentences were incorporated for more comprehensive scenarios. This shares a similar idea to our work, yet the challenge does not give multi-sentence materials at scale for detailed study. 

Recently, the most widely used NLI benchmarks include the Stanford Natural Language Inference (SNLI) dataset \cite{snli:emnlp2015}, and the subsequently expanded MultiNLI \cite{williams-etal-2018-broad}, bringing sentences of various genres into the original SNLI. MultiNLI is included in the GLUE benchmark \cite{DBLP:journals/corr/abs-1804-07461} and is widely used in evaluating language models' performance. 
Other NLI datasets include the Question-answering NLI (QNLI) \cite{DBLP:journals/corr/abs-1804-07461}, 
the Winograd NLI (WNLI) \cite{DBLP:journals/corr/abs-1804-07461}, 
the SciTail \cite{Khot2018SciTaiLAT} etc., which focuses on different aspects of knowledge. While all the above datasets are on the sentence-level, we investigate NLI for long texts.
%The GLUE benchmark also introduces inference tasks such as The Stanford Question Answering Dataset (QNLI) \cite{Rajpurkar_2016} and The Winograd Schema Challenge (WNLI) \cite{10.5555/3031843.3031909}, which are a two-way classification of sentence pairs. 

Dialogue NLI \cite{welleck2018dialogue} features a persona-based dialogue structure for making inference on the current utterance based on previous dialogue history. Similar to our dataset, discourses involve multi-sentence context as premises. However, they do not consider relationships that require more than two sentences to express, nor is logical reasoning explored.
%SciTail \cite{Khot2018SciTaiLAT} is a science-based entailment benchmark, which consists of premise-hypothesis sentence pairs adapted from science questions into a 2-way decision task. The contexts of scientific questions are limited to sentence-level, which is contrast to our dataset that requires multi-sentence reasoning.

Adversarial NLI \cite{Nie2019AdversarialNA} introduces an iterative, adversarial human-and-model-in-the-loop training method to collect large-scale NLI dataset. It holds the simple intuition that longer contexts lead to harder examples, which coincide with our idea to some extent.  The Adversarial NLI dataset is similar to ours in that longer contexts are considered in the premises. However, we differ in context length and reasoning types. The context of our dataset is much longer and with \textit{multiple} paragraphs being involved. In contrast, Adversarial NLI has single-paragraph contexts only. In addition, it does not test logical reasoning, which is the main focus of ConTRoL. To our knowledge, we are the first to introduce a passage-level NLI dataset requiring comprehensive grounded logical reasoning.

AlphaNLI \cite{ch2019abductive} explores the problem of abductive reasoning. It asks for the most plausible explanation given observations from \textit{two} narrative contexts. Similar to our dataset, investigating the nature of human reasoning is the target of AlphaNLI. However, the AlphaNLI challenge resembles the classical formulation of abductive reasoning, which is different from the reasoning types we are focused on. In addition, both premises (i.e., observation contexts) in AlphaNLI are written in single sentences. In contrast, our dataset consists of multi-paragraph premises.

\citet{clark2020transformers} investigated deductive reasoning by synthesizing a dataset related to NLI. The input is a set of facts and a set of rules that explain the facts, which can be viewed as a premise, together with a fact that can be viewed as a hypothesis, and the output is a binary class {\it true} or {\it false}. Compared with their dataset, our work differs in two aspects. First, we consider more reasoning types. Second, ConTRoL is a multi-paragraph NLI dataset with human-written inputs.
\textbf{Contextual Reasoning}
Long texts with multiple paragraphs have been explored in reading comprehension. In particular, there have been challenges that examine evidence integration over multiple text passages \cite{welbl2017constructing,Yang_2018,squad}, and challenges that focus on commonsense reasoning \cite{commonsenseqa,mutual}, including social commonsense \cite{social-iqa} and external knowledge \cite{cosmos,winogrande}. Different from these datasets, ConTRoL examines more complex contextual reasoning types such as {\it logical reasoning}.

There have been reading comprehension datasets that examine logical reasoning. LogiQA \cite{Liu_2020} is sourced from public service exams. It focuses on linguistic reasoning questions typically featured with a question and four possible answers. ReClor \cite{yu2020reclor} is a reading comprehension dataset that is sourced from the GMAT and LSAT test. Similar to our dataset, these datasets examine a range of different logical reasoning types. Different from these benchmarks, ConTRoL takes the form of NLI, which is a more fundamental linguistic task and relevant to different downstream tasks. The correlation and differences between exisiting datasets are shown in Table \ref{Tab:benchmarks}.

\begin{table}
\centering
\begin{tabular}{c|c}
\hline
 & \textbf{ConTRoL} \\
\hline
Construction Method & Exams \\
Context Type & Passage \\
\hline
\# of passages & 1,970 \\
\# of premise-hypothesis pairs & 8,325 \\
\# of multi-paragraph & 4,171 \\
Avg. length of multi-paragraph & 757 \\
\# of single-paragraph & 4,154 \\
Avg. length of single-paragraph & 148 \\
\hline
Vocab size (premise) & 54,265 \\
Vocab size (hypothesis) & 14,323 \\
Avg. premise length  & 452 \\
Avg. hypothesis length & 12 \\
\hline
Lexical overlap (Entailment) & 4.87\% \\
Lexical overlap (Neutral) & 4.19\% \\
Lexical overlap (Contradiction) & 5.49\% \\
\hline
\end{tabular}
\caption{Data statistics of ConTRoL.}
\label{tab:stat}
\end{table}

\section{Dataset}

Crowdsourcing has been a widely-adopted practice for developing large-scale NLI datasets \cite{snli:emnlp2015, williams-etal-2018-broad, Nie2019AdversarialNA}. However, producing a high-quality dataset addressing complex logical reasoning can be difficult for crowdsource workers. Annotation artefacts exist in crowdsourced datasets, for the annotation protocols encourage workers to adopt heuristics to generate hypotheses quickly and efficiently \cite{DBLP:journals/corr/abs-1803-02324}. 
%For example, common strategies are illustrated in the SNLI annotation protocol: entailed hypotheses are generated by removing gender or number information; neutral hypotheses can be constructed by adding a purpose clause and contradictions are associated with negations.
To avoid such issues, we source our dataset from examinations, and in particular senior aptitude tests (verbal reasoning test), which are designed by experts. 

\subsection{Data Collection and Statistics}
We collect our data from publicly available online practice tests, which include verbal logical reasoning tests in the Police Initial Recruitment Test (PIRT), verbal reasoning tests used by the Medical College Admission Test (MCAT) and University Clinical Aptitude Test (UCAT), as well as verbal aptitude tests adopted by corporations' employee recruitment \& selection online test. Unlike reading comprehension tests, which can be diverse both in question types and options, questions in the original verbal reasoning tests are similar in struct to NLI tests, where a premise and a hypothesis are given, and the answer is a choice from three options: \textit{true}, \textit{false} and \textit{cannot say}. This corresponds to the three-label setting of the NLI task and we can easily convert the three answer choices into \textsc{entailment}, \textsc{contradiction} and \textsc{neutral} respectively.

The verbal reasoning tests require exam-takers to comprehend meaning and significance, assess logical strength, make valid inference, and identify a valid summary, interpretation or conclusion. The subjects of the passages are drawn from a range of fields, such as current affairs, business, science, the environment, economics, history, meteorology, health and education. The questions are of high quality,  advanced in difficulty level, used in exams such as police initial selection and other highly intellectual practices' candidate recruitment.

The detailed statistics of ConTRoL are shown in Table~\ref{tab:stat}. After removing all duplicated questions, we obtain 8,325 context-hypothesis pairs. We also calculate the lexical overlap between context and hypothesis, finding only 4.87\% overlap in the \textsc{Entailment} relationship, and 5.49\% in the \textsc{Contradiction} relationship. This suggests that ConTRoL can be difficult to solve by plain lexical matching.

\begin{table*}
\centering
\begin{tabular}{c|c|c|c|c|c|c|c|c|c|c|c}
\hline
 & \multicolumn{2}{c|}{\textbf{Overall}} & \multicolumn{3}{c|}{\textbf{Entailment}} & \multicolumn{3}{c|}{\textbf{Neutral}} & \multicolumn{3}{c}{\textbf{Contradiction}}\\
\hline
& Acc & Avg.F1 & Precision & Recall & F1 & Precision & Recall & F1 & Precision & Recall & F1 \\
\hline
Human & 87.06 &93.15  & 94.83 & 95.65 & 95.24 & 93.33 & 91.21 & 92.26 & 93.02 & 90.91 & 91.95 \\
Ceiling & 94.40 & 97.26 & 99.16 & 99.16 & 99.16 & 97.72 & 93.75 & 95.69 & 96.09 & 97.79 & 96.93 \\
\hline
BERT-base & 47.39 & 46.22 & 43.84 & 54.40 & 42.45 & 39.67 & 51.07 & 50.21 & 41.65 & 52.68 & 46.00 \\
BERT-large & 50.62 & 49.49 & 45.15 & 59.32 & 45.96 & 44.21 & 53.52 & 53.19 & 44.68 & 56.27 & 49.31 \\
RoBERTa & 45.90 & 45.67 & 40.99 & 51.24 & 45.38 & \textbf{47.93} & 44.34 & 45.96 & 44.19 & 47.54 & 45.67  \\
Longformer & 49.88 & 46.22 & 43.24 & 58.88 & 45.64 & 46.28 & 54.74 & 46.81 & 44.71 & 56.74 & 46.22   \\
XLNet & 54.85 & \textbf{54.93} & 46.15 & 62.22 & \textbf{54.13} & 47.11 & 59.94 & 55.74 & 46.63 & 61.06 & \textbf{54.93} \\
BART & \textbf{56.34} & 54.18 & \textbf{50.23} & \textbf{67.32} & 49.12 & 44.21 & \textbf{62.99} & \textbf{59.57} & \textbf{47.03} & \textbf{65.09} & 53.85 \\
\hline
BART-NLI & 45.02 & 42.33 & 39.85 & 53.49 & 40.87 & 43.80 & 46.79 & 43.83 & 41.73 & 49.92 & 42.30 \\
BART-NLI-FT & 60.95 & 57.41 & 62.58 & 61.54 & 58.67 & 42.15 & 78.29 & 56.17 & 50.37 & 68.91 & 57.39 \\
\hline
\end{tabular}
\caption{\label{Tab:results}
Experiment results on ConTRoL. BART-NLI indicates training on SNLI, MultiNLI and Adversarial NLI and testing on ConTRoL. BART-NLI-FT indicates BART-NLI followed by a fine-tuning step on ConTRoL.
}
\end{table*}

\subsection{Data Format}
The data format of ConTRoL follows existing NLI benchmarks \cite{snli:emnlp2015, williams-etal-2018-broad}, where each instance contains a premise, a hypothesis, and a label from \textsc{Entailment}, \textsc{Neutral} and \textsc{Contradiction}. 
Different from existing datasets, the premises are much longer, in one or more paragraphs. In addition, for each premise, three or more hypotheses are given, which is another distinction from former NLI datasets.

%Different with previous NLI benchmarks \cite{}, ConTRoL 

\subsection{Reasoning Types}
We manually categorize the test instances by the reasoning type, which can be described as follows:
\begin{itemize}

    \item \textbf{Coreferential Reasoning over Long Texts}

Coreferential reasoning \cite{ye2020coreferential} is a form of reasoning over multiple mentions. Long text can accommodate complex relationships between noun phrases,  which makes coreferential reasoning crucial for the coherent understanding of texts.

   \item \textbf{Verbal Logical Reasoning}

Verbal logical reasoning \cite{Liu_2020} is the ability to examine, analyze, and critically evaluate arguments as they occur in ordinary language.
In contrast to formal logical reasoning, most of which uses abstract diagrammatical cues, verbal logical reasoning concerns the logical inference of human language. Deep logical reasoning can be necessary in comprehending long texts. 

    \item \textbf{Temporal and Mathematical Reasoning}

Time and sequential cues of events and requires the ability to reason about time and do the necessary mathematical calculation. Temporal reasoning \cite{10.3115/976858.976900} is the process of extracting temporal cues and combining them into a coherent temporal view. Various types of temporal information can be found in ConTRoL.

    \item \textbf{Information Integration over Paragraphs}
    
Multi-step reasoning \cite{liu2020multistep,chen2019multihop,welbl2017constructing} is the ability to retrieve and combine information from multiple paragraphs or multiple documents. For each hypothesis, readers find the most relevant paragraphs in a premise through an iterative (multi-step) process between the contexts and the hypotheses.

\item \textbf{Analytical Reasoning}

Analytical reasoning \cite{WILLIAMS2019574} is the ability for problem solving to consider a group of facts and rules, and determine the validity of new facts. The fact sets are based on a single or multiple paragraphs, reflecting the kinds of detailed analyses of relationships and sets of constraints. Reasoning is based on what is required given the scenario, what is permissible given the scenario, and what is prohibited given the scenario.
\end{itemize}

Examples of the above reasoning types can be found in Figure \ref{fig:types}. Among all the reasoning types, logical reasoning takes 36.2\% of all the test instances, followed by information integration, which takes 32.6\%. The proportion of coreferential reasoning, analytical reasoning and temporal reasoning are 26.0\%, 12.8\% and 12.4\%, respectively. It is also worth noticing that one context-hypothesis pair may contain more than one reasoning type, under which circumstance we take the most significant one into the statistics.

\section{Models}
We establish several strong baseline methods using the state-of-the-art pre-trained language models.
% \subsection{Random and BOW overlap}
% Random guess are provided as the lower bound of model performance. We compute BOW (bag-of-word) overlap to ensure that the CTRL benchmark cannot be solved by trivial means. BOW model explores bag-of-word features of both the context and the hypothesis and calculate overlap. The simple baseline results represent the statistical expectation of our dataset. 

\subsection{Pre-trained Language Models}
% Rather than manually specify all the features, deep neural methods use various neural network architectures to discover useful features in data through training. 
% Pre-trained language models achieve great success on various NLP tasks including NLI.

% We establish several strong baseline using state-of-the-art pre-trained language models. 

% \textbf{GPT}
% GPT \cite{} is an autoregressive transformer-based language model which can be deployed to all tasks without gradient update or fine-tuning.

\textbf{BERT}
\cite{devlin-etal-2019-bert} is a Transformer-based \cite{vaswani2017attention} language model. During pre-training, BERT uses a masked language modeling objective. The basic idea is to train a model to make use of bidirectional context information for predicting a masked token, so that linguistic knowledge can be collected from large texts. It has been shown that such a language model contains certain degrees of syntactic \cite{syntactic}, semantic \cite{semantic}, commonsense \cite{commonsense} and logical reasoning \cite{clark2020transformers} knowledge.
%Following , we concatenate the context and the hypothesis together as input, conduct a three-way classification both on pre-trained bert-base and bert-large model. We report the best performance after fine-tuning.

\begin{figure}[t]
\centering
\includegraphics[width=0.3\textwidth]{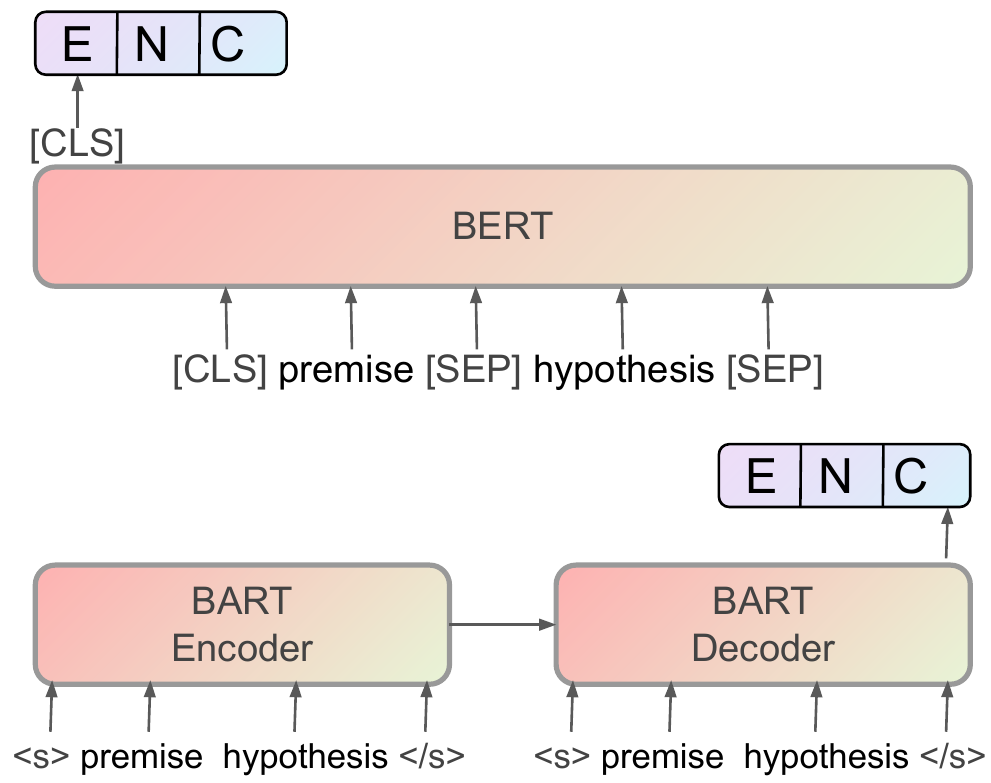} % Reduce the figure size so that it is slightly narrower than the column.
\caption{The model structure of BERT and BART. (``E" represents \textsc{Entailment}, ``N" represents \textsc{Neutral}, ``C" represents \textsc{Contradiction}}
\label{fig:models}
\end{figure}

\textbf{RoBERTa}
\cite{liu2019roberta} extends BERT using a more dynamic sentence masking method.
% It is not the case when testing on the CTRL dataset. RoBERTa's performance is slightly lower than that of BERT, indicating the weakness in dealing with more complex contexts. 

\textbf{XLNet} \cite{yang2019xlnet}
Instead of a bidirectional Transfomer, XLNet uses Transformer-XL \cite{Dai_2019} as its main structure. Avoiding several limitations faced by BERT, XLNet uses an autogressive language model which models long-term dependencies beyond a fixed context.

\textbf{Longformer} \cite{beltagy2020longformer}
Traditional self-attention operation are unable to process long sequences, which scales quadratically with the sequence length.
The aforementioned Transformer-based models constrain the input to 512 tokens. To address this limitation, Longformer adopts sliding window attention with global attention to replace the self-attention mechanism in pretrained Transformers.

\textbf{BART} \cite{Lewis_2020} is a denoising autoencoder
for pre-training sequence-to-sequence models by combining bidirectional and auto-regressive Transformers.
% Unlike previous NLI benchmarks, CTRL is a passage-level NLI dataset with an amounts of contexts exceeding the maximum token limit of transformer models. This represents a big challenge for current SOTA approaches. To tackle this problem, Longformer\cite{beltagy2020longformer} break the token limit with attention mechanism that scales linearly with sequence length. 
% Following \citeauthor{beltagy2020longformer}, we test Longformer's performance on the CTRL dataset.
\subsection{NLI Model}
The NLI model structures of BART and BERT-based are illustrated in Figure \ref{fig:models}.
For BERT-based models (i.e., BERT, RoBERTa, XLNet and Longformer), following \citet{devlin-etal-2019-bert}, given a premise $P$ and a hypothesis $h$, we concatenate premise-hypothesis pair as a new sequence $[CLS] + p + [SEP] + h + [SEP]$, where [CLS] and [SEP] are special symbol for classification token and separator token. After pre-training model encoding, the last layer's hidden representation from the [CLS] token is fed in an MLP+softmax for classification. 
For BART, we feed the same sequence to both the encoder and the decoder, using the last hidden state for classification.
The class that corresponds to the highest probability is chosen as model prediction. 

\subsection{Implementation Details}

We randomly split the dataset into training, development, and test set with the ratio of 8:1:1. 
All models are trained for 10 epochs. We find hyper-parameters using grid search: batch size $\in \{8, 16, 32\}$ learning rate $\in \{1e-5, 2e-5, 3e-5, 4e-5, 5e-5\}$ and gradient accumulate step $\in \{1, 2, 4\}$.  We set the max length to 512 tokens for all models except Longformer, of which 3,000 tokens are the max length we take. Models with the best performance on the development set are used for testing.

\subsection{Evaluation}
Following the NLI benchmark setting  \cite{snli:emnlp2015,williams-etal-2018-broad,welleck2018dialogue}, we employ the overall accuracies as the main evaluation method. Furthermore, to give more detailed analysis, we also calculate precision, recall and F1 score on the \textsc{entailment}, \textsc{neutral} and \textsc{contradiction} labels.
% Different evaluation metrics are used to showcase the result. 
% Exact-matching accuracy gives the intuitive understanding, while the F-measure, combines both precision and recall into the calculation, provide a balanced representation of our measurement. 
% Apart from overall measurement, we also look into the performance on the three labels individually. Precision,recall and F1 score are calculated on the label set of entailment, neutral and contradiction.
\begin{table*}
\centering
\begin{tabular}{c|r|r|c|c|c|c}
\hline
\textbf{Benchmark} & \textbf{\# Train} & \textbf{\# Test} & \textbf{BERT}  & \textbf{SOTA Model} & \textbf{SOTA Performance} & \textbf{Human} \\
\hline
%SNLI & 90.0 & & & SemBERT \cite{Zhang_2020} & 91.9 & 94.1\\
MultiNLI & 393k & 20k &85.9 & T5-11B \cite{raffel2019exploring} & 92.0 & 92.8\\
QNLI & 105k & 5.4k & 92.7  & ALBERT \cite{lan2019albert} & 99.2 & 91.2 \\
RTE  & 2.5k & 3k &70.1  & T5-11B \cite{raffel2019exploring} & 92.5 & 93.6 \\
WNLI & 634 & 146 &65.1  & T5-11B \cite{raffel2019exploring} & 93.2  & 95.9 \\
\hline
ConTRoL & 8.3k & 804 & 50.6  & BART-NLI-FT & 61.0 & 94.4\\
\hline
\end{tabular}
\caption{\label{Tab:sota}
The state-of-the-art performances of popular NLI benchmarks (accuracy\%).
}
\end{table*}

\subsection{Human Performance}
To measure human performance on the ConTRoL dataset, we randomly select 300 context-hypothesis pairs from the test set. Four testees are recruited. The testees are well educated, two of them are post-graduate students and two of them have PhD degrees. We report the human performance by the mean score and standard deviation. The human ceiling performance is obtained by considering the proportion of questions with at least one correct answer.

\section{Results}

Table \ref{Tab:results} shows the main results. 
As shown in the table, BERT gives an overall accuracy of 50.62\% and F1 of 49.49\%; RoBERTa gives a higher accuracy of 45.90\% and F1 of 45.67\%; Longformer gives an overall accuracy of 49.88\% and F1 of 46.22\%;
XLNet gives an overall accuracy of 54.85\% and F1 of 54.93\%.
The top reported performance is given by the BART model, with a 56.34\% accuracy score. Compared with human performance, the performance of BART is lower by approximately 30\%.
The human performance on the ConTRoL surpasses SOTA NLI models by a large margin, which demonstrate limitations for the computational models for solving contextual reasoning tasks.

%We test BERT-base and BERT-large to examine the effect of parameter size.
%The RoBERTa model is outperformed by the BERT model on our dataset, a similar report is seen in the experiment of Adversarial NLI \cite{Nie2019AdversarialNA}. The two cases show that RoBERTa generally struggles with long contexts. 
%The performance of Longformer is similar to that of BERT-large, we do not see much increase of accuracy on the Longformer model. 

\begin{figure}[t]
\centering
\includegraphics[width=0.4\textwidth]{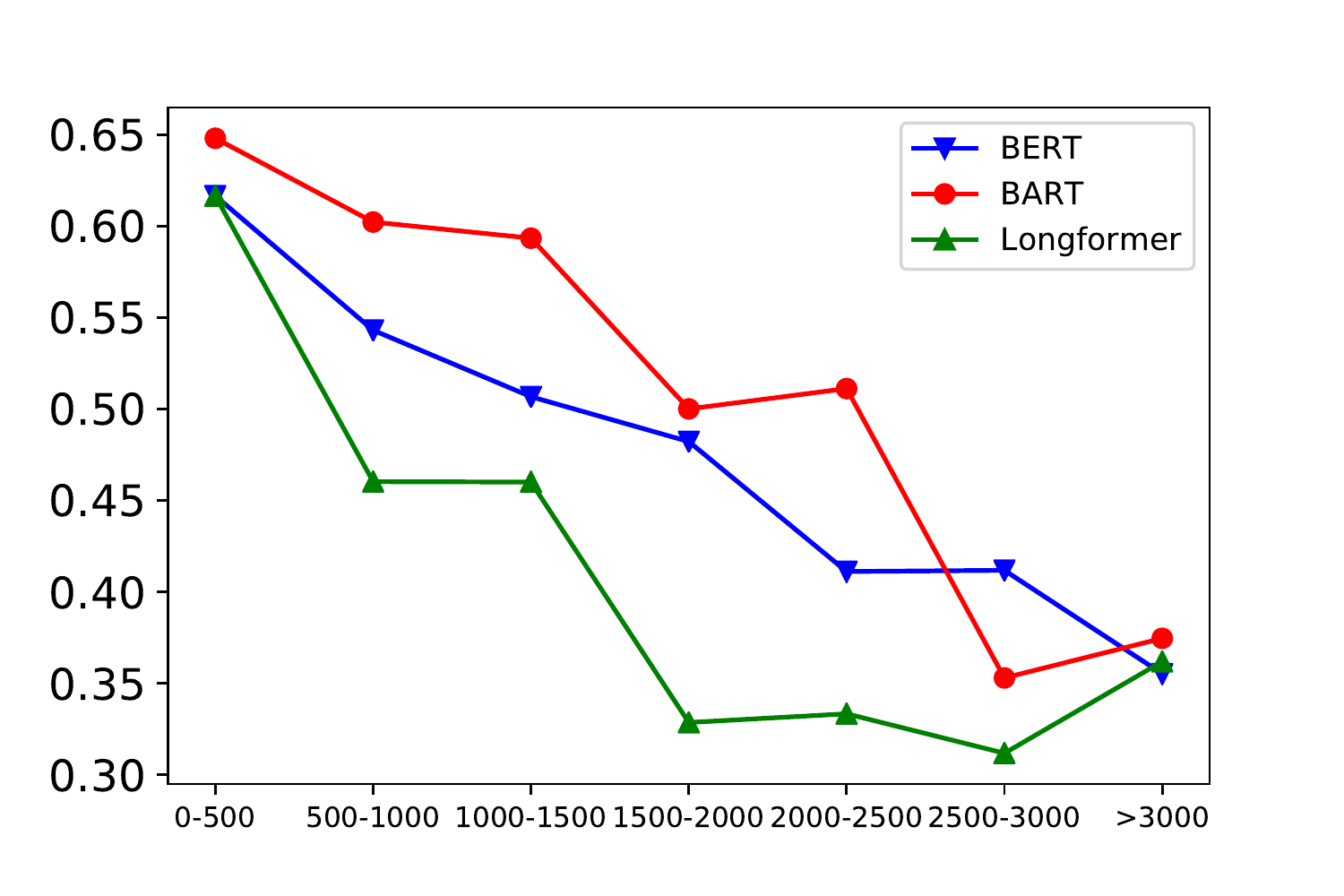} % Reduce the figure size so that it is slightly narrower than the column.
\caption{Performance across different context lengths.}
\label{fig:length}
\end{figure}

\begin{table}
\centering
\begin{tabular}{c|cc}
\hline
\textbf{Reasoning Type} & \textbf{BERT} & \textbf{BART} \\
\hline
Coreferential Reasoning  & 74.64 & 74.92\\
Analytical Reasoning  & 67.96 & 69.65\\
Temporal Reasoning  & 56.44 & 57.34\\
Information Integration & 40.07 & 43.39\\
Logical Reasoning  & 40.76 & 43.20\\
\hline
\end{tabular}
\caption{\label{across-types}
Performance across reasoning types (accuracy\%).
}
\end{table}

As shown in Table \ref{Tab:sota}, we see a huge performance drop when the SOTA model results on ConTRoL are compared to their reported score on previous NLI dataset \cite{liu2019roberta}.
In contrast, similar to the existing benchmarks, human testees are able to achieve high scores with proper training. Different from datasets that emphasise fact extraction and verification, the inference of ConTRoL relies not only on the long-term dependency of texts, but the contextual reasoning abilities regarding long contexts. 
%Humans can still achieve high performance on the ConTRoL dataset. Because human testees usually look at the context and hypothesis back and forth when deciding the relationship. 

To further understand the phenomena, we conduct various qualitative and quantitative detailed analysis on ConTRoL.

\subsection{Performance Across Different Relationships}
We first compare human performance and model performance across different relationships.
% we observe some interesting phenomena when comparing human performance to model performance on different labels.
Interestingly, as shown in Table \ref{Tab:results}, humans are good at deciding the entailment and contradiction relationship, while struggling when examining the relationship of neutral. This can be because humans tend to associate external irrelevant knowledge to the reasoning process, which is not expressed in the context. The computational models seem not to bear this burden, which gives similar results across the three labels.

%\section{Discussion}

\subsection{Performance Across Different Context Lengths}
 As mentioned earlier, aside from single-paragraph context-hypothesis pairs, there are multi-paragraph context-hypothesis pairs in our dataset. We conduct experiments on the single-paragraph and multi-paragraph instances separately, which gives us the insight into how context length affects the performance of the transformer-based NLI models. The accuracy of the BERT model is 40.30\% on multiple-paragraph instances while 51.17\% on single-paragraph instances. 
 We also conduct fine-grained analysis concerning the context length. The result are shown in Figure \ref{fig:length}. When the context length increases, the model performance drops accordingly. The best model BART drops from 65\% (shorter than 500 words) to 40\% (longer than 3,000 words), demonstrating that ConTRoL heavily rely on passage-level reasoning ability, rather than sentence-level reasoning ability.

\subsection{Performance Across Reasoning Types}
Table \ref{across-types} gives the performance over the 5 reasoning types. BERT and BART have similar trends across different reasoning types. In particular, on the coreferential reasoning type, BERT and BART give accuracies of 74.64\% and 74.92\%, respectively. On the other hand, both models are more confused on reasoning types including multi-step reasoning and logical reasoning. This can be because multi-step reasoning can be correlated with longer context length, and information integration is processed over multiple paragraphs. Finally performing inductive and deductive reasoning is difficult for current models, making logical reasoning a difficult endeavour \cite{Liu_2020}.

\subsection{Transfer Learning}
Recent studies have shown the benefit of fine-tuning on similar datasets for knowledge transfer \cite{cosmos}. We explore three related NLI datasets for knowledge transfer, SNLI~\cite{snli:emnlp2015}, MultiNLI~\cite{williams-etal-2018-broad} and Adversarial NLI \cite{Nie2019AdversarialNA}. As shown in the last two rows of Table~\ref{Tab:results}, BART-NLI only achieves 45.0\%, which shows that ConTRoL is different from existing NLI benchmarks. After fine-tuning on ConTRoL, BART-NLI-FT achieves the state-of-the-art results, which demonstrates that general knowledge from traditional NLI benchmarks are beneficial to the performance of ConTRoL.

\section{Discussion}

\subsection{Corpus Bias}
\begin{figure}[t]
\centering
\includegraphics[width=0.4\textwidth]{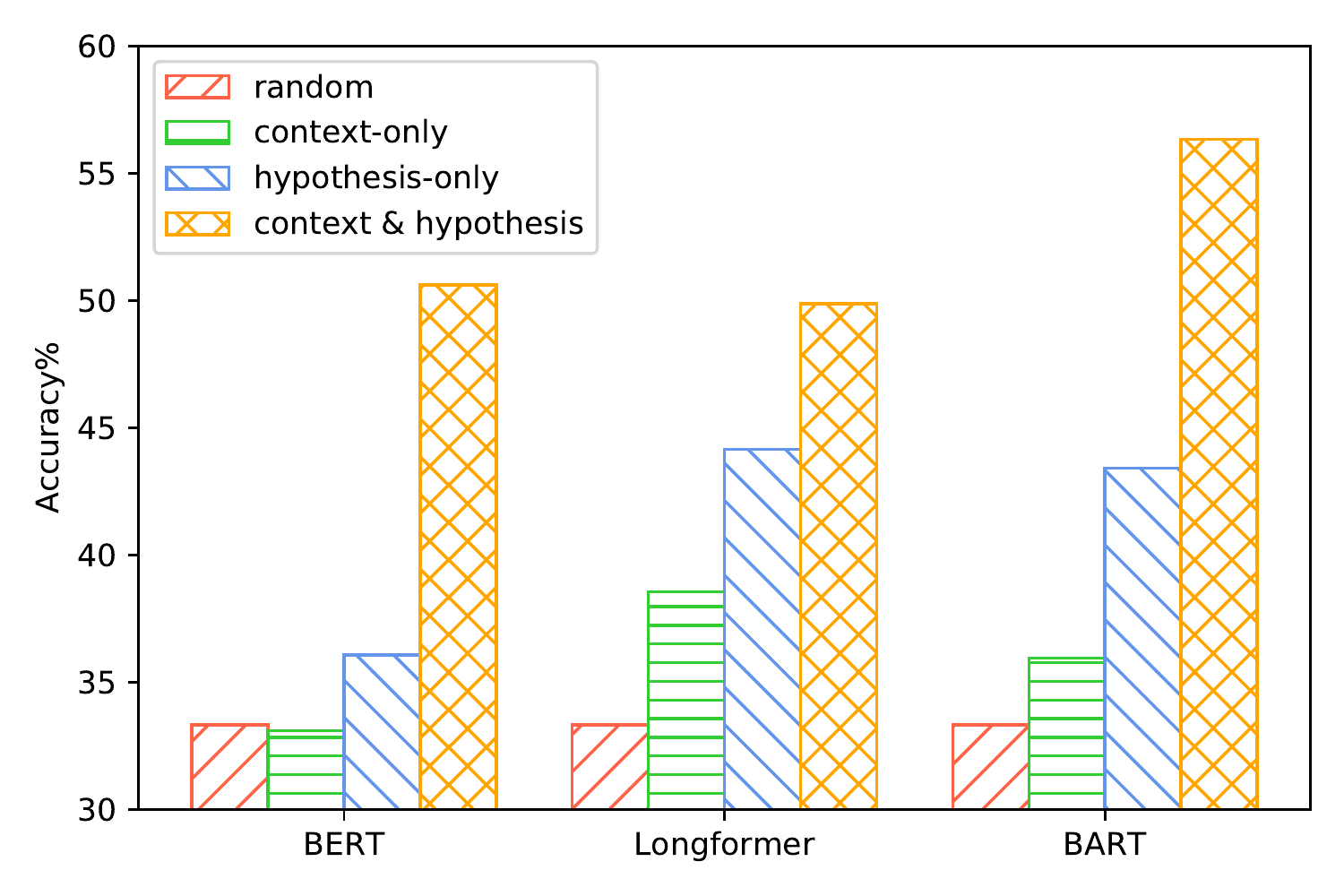} % Reduce the figure size so that it is slightly narrower than the column.
\caption{Ablation study on different models.}
\label{fig:ablation}
\end{figure}

%\textbf{Hypothesis only}
Recent studies show that pre-trained language models can make the right prediction by merely looking at context \cite{mccoy-etal-2019-right}. The hypothesis-only bias is common in large-scale datasets for NLI, particularly for benchmarks constructed by crowdsourcing methods. We conduct an ablation experiment on ConTRoL. Figure \ref{fig:ablation} shows the comparison of BERT, Longfomer and BART. BERT gives a 36.07\% accuracy with hypothesis-only, which is slightly higher than theoretical random guess; Longformer gives a 44.15\% accuracy, surpass BART by a small margin, which gives a 43.41\% accuracy. 

Context-only results are also calculated to further examine annotation artefacts in the ConTRoL dataset. BERT gives 33.09\% accuracy; BART gives 35.94\% accuracy; Longformer also gives a better performance than BERT and BART, which gives 38.56\% accuracy. Longformer gives a better score on context-only and hypothesis-only ablation, which can be because Longformer sees more context than the other two models. The ablation results are lower than the results without ablation, which indicates that models need to look at both the contexts and the hypotheses to make the correct prediction. Thus we conclude that the ConTRoL dataset is exempt from significant annotation biases thanks to expert-designed questions.

%%We train a RoBERTa model on the datasets of SNLI, MNLI, and Adversarial NLI all together. Then we test the model on our test set.

\subsection{Case Study}
\begin{figure}[t]
\centering
\includegraphics[width=0.45\textwidth]{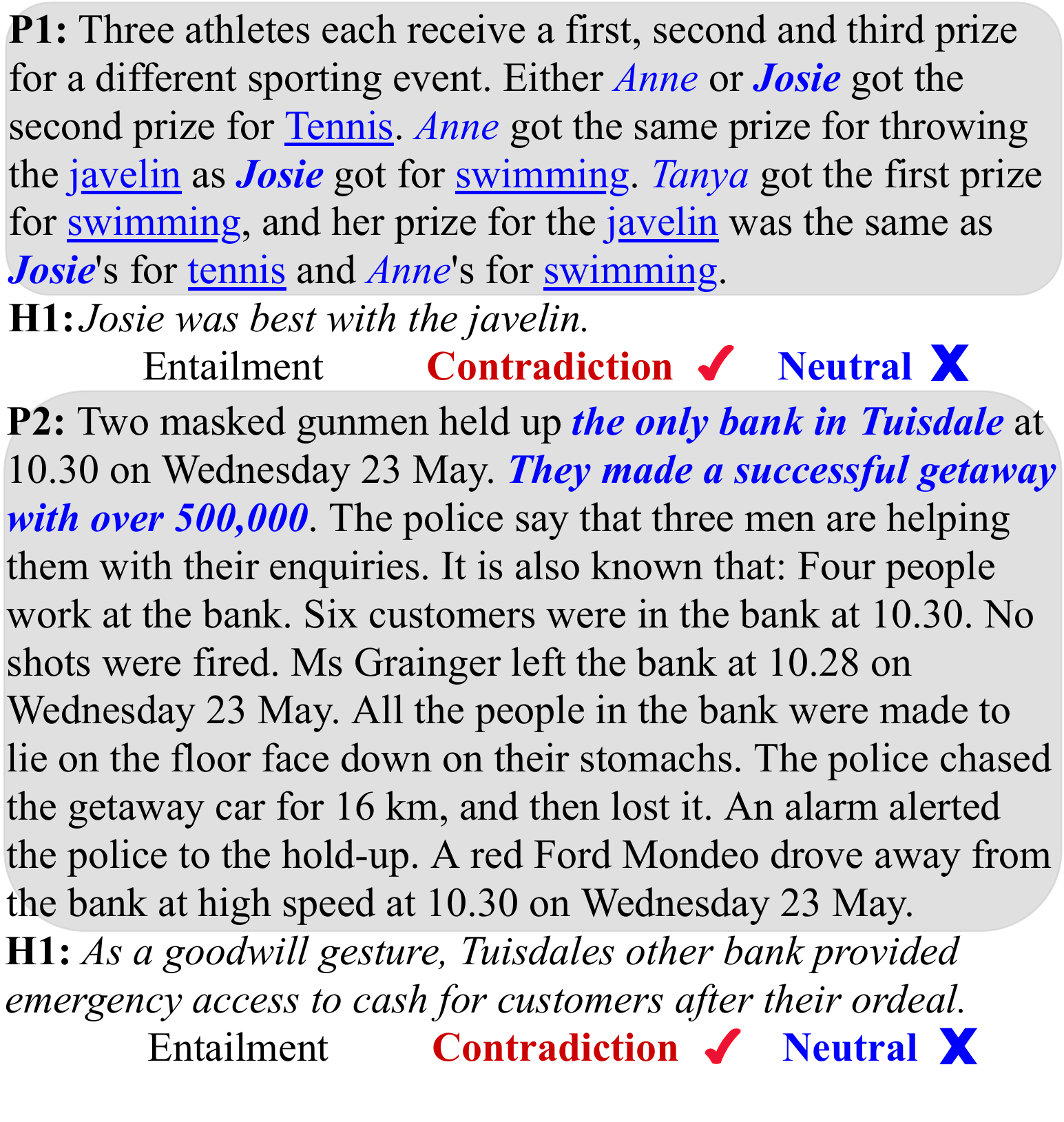} % Reduce the figure size so that it is slightly narrower than the column.
\caption{Example mistakes of BART ({\color{red} \cmark} indicates the correct label and {\color{blue} \xmark} indicates the BART prediction. Reasoning clues are highlighted in the context.)}
\label{fig:case}
\end{figure}
Figure \ref{fig:case} shows two cases that demonstrate the challenge in ConTRoL. P1 of Figure \ref{fig:case} is a representative example of the challenges brought by logical reasoning. The context concerns three athletes and three sports. We need to decide their places in a competition. The lexical overlap between the premise and the hypothesis is very low. BART incorrectly chooses the \textit{Neutral} label, while we can infer from the context that Josie is actually not the best with the javelin, which can only be done by deductive reasoning.
Information integration is difficult for BART. 

P2 of Figure \ref{fig:case} shows a typical example of challenge brought by information integration, where the hypothesis is made considering the whole passage. We know from the first sentence that Tuisdale holds the only bank in the region. The hypothesis talks about the possible aftermath of the robbery, BART incorrectly chooses the \textit{Neutral} label for it overlooks the information that Tuisdale only has one bank.In both cases, the correct answer is not explicitly mentioned in the premise, but need contextual reasoning to infer.

\section{Conclusion}
We presented the ConTRoL dataset, a passage-level NLI benchmark that consists of different contextual reasoning types. Compared with existing NLI benchmarks, the context length of the premise is bigger by a large margin, and reasoning skills such as logical reasoning, analytical reasoning and multi-step reasoning are required. Experiments show that state-of-the-art NLI models perform poorly on the ConTRoL dataset, far below human performance. Ablation study indicates that the data does not suffer from heavy annotation artefacts and can be served as a reliable NLI benchmark for future study. To our knowledge, we are the first to introduce a passage-level NLI dataset that highlights contextual reasoning.

% References and End of Paper
% These lines must be placed at the end of your paper
\bibliography{aaai.bib}
\end{document}